\documentclass[conference]{IEEEtran}

\IEEEoverridecommandlockouts

\usepackage{algorithmic}
\usepackage{amsmath,amssymb,amsfonts}
\usepackage{cite}
\usepackage{graphicx}
\PassOptionsToPackage{hyphens}{url}\usepackage{hyperref}
\usepackage{subcaption}
\usepackage{textcomp}
\usepackage{tikz}
\usepackage{xcolor}

\def\BibTeX{{
  \rm B\kern-.05em{\sc i\kern-.025em b}\kern-.08em
  T\kern-.1667em\lower.7ex\hbox{E}\kern-.125emX
}}

\begin{document}

\title{
  xYOLO: A Model For Real-Time Object Detection In Humanoid Soccer On Low-End
  Hardware
  \thanks{
    This project would like to thank the following departments for funding:
    Computer Science, HIT Lab NZ and School of Product Design.
  }
}

\author{
  \IEEEauthorblockN{
    Daniel Barry \IEEEauthorrefmark{1},
    Munir Shah \IEEEauthorrefmark{5},
    Merel Keijsers \IEEEauthorrefmark{2},
    \\
    Humayun Khan \IEEEauthorrefmark{4} and
    Banon Hopman \IEEEauthorrefmark{3}
  }
  \\
  \IEEEauthorblockA{
    \textbf{Departments:}
    \IEEEauthorrefmark{1}
      \textit{Computer Science}
    \IEEEauthorrefmark{2}
    \IEEEauthorrefmark{4}
      \textit{HIT Lab NZ}
    \IEEEauthorrefmark{3}
      \textit{School of Product Design}
  }
  \IEEEauthorblockA{
    \textbf{Emails:}
    \IEEEauthorrefmark{1}
      dan.barry@pg.canterbury.ac.nz
    \IEEEauthorrefmark{5}
      munirsha@gmail.com
    \\
    \IEEEauthorrefmark{2}
      merel.keijsers@pg.canterbury.ac.nz
    \IEEEauthorrefmark{3}
      bmh77@uclive.ac.nz
    \IEEEauthorrefmark{4}
      humayun.khan@pg.canterbury.ac.nz
  }
  \\
  \IEEEauthorblockA{
    \textit{University of Canterbury} \\
    Christchurch, New Zealand
  }
}

\maketitle

\begin{abstract}
  With the emergence of onboard vision processing for areas such as the
  internet of things (IoT), edge computing and autonomous robots, there is
  increasing demand for computationally efficient convolutional neural network
  (CNN) models to perform real-time object detection on resource constraints
  hardware devices. Tiny-YOLO is generally considered as one of the faster
  object detectors for low-end devices and is the basis for our work. Our
  experiments on this network have shown that Tiny-YOLO can achieve 0.14 frames
  per second (FPS) on the Raspberry Pi 3 B, which is too slow for soccer
  playing autonomous humanoid robots detecting goal and ball objects. In this
  paper we propose an adaptation to the YOLO CNN model named xYOLO, that can
  achieve object detection at a speed of 9.66 FPS on the Raspberry Pi 3 B. This
  is achieved by trading an acceptable amount of accuracy, making the network
  approximately 70 times faster than Tiny-YOLO. Greater inference speed-ups
  were also achieved on a desktop CPU and GPU. Additionally we contribute an
  annotated Darknet dataset for goal and ball detection.
\end{abstract}


\section{Introduction}
\label{sec:introduction}

  The popularization of deep learning has done much to further advancements in
  computer vision, where modest amounts of computational power allow for the
  processing of images to gain insight on their contents
  \cite{LenCunBenginoHinton2015}. As real-world image data typically has high
  spatial correlation, convolutional neural networks (CNNs) have been
  particularly successful in the application of object detection in images
  \cite{ZellerFergus2014}. Compared to fully connected networks, CNNs offered
  large computation and size reduction \cite{HochreiterBengioFrasconiSchmidhuber2001}.

  In this paper our chosen problem domain is RoboCup, an annually held
  international humanoid robotics competition with the goal of producing a team
  of robots that beat the best human football players in the world by 2050
  \cite{KitanoAsadaKuniyoshiNodaOsawa1997}. The reason for this ambitious goal
  is to help motivate several key areas in artificial intelligence in a format
  that the general public understand and appreciate without having insight to
  the complexity of the robots themselves. There are many limitations in the
  competition in aid of reaching the 2050 goal, such as: robots must be fully
  autonomous, must be human like (e.g. passive sensors only) and must adhere to
  an adaptation of the official FIFA rules \footnote{Humanoid rules:
  \url{https://www.robocuphumanoid.org/materials/rules/}}.

  Our team, \emph{Electric Sheep}, competed in the 2019 world cup with our
  unique low-cost open-source humanoid robotics platform \footnote{Humanoid
  platform: \url{https://github.com/electric-sheep-uc/black-sheep}}. The
  purpose of designing and building a low-cost platform was to lower the
  boundary to entry, as larger robots have seen larger costs in recent years
  which could discourage new teams from entering the competition. Our platform,
  \emph{Black Sheep}, performs all processing on a Raspberry Pi 3 B including
  our vision processing pipeline. As our agent behaviour is very simple it only
  requires the detection of goals and balls.

  In this paper, we propose a YOLO based CNN model which can detect balls and
  goal posts at $\approx$10 FPS, which given the current relatively slow speed
  of robot play, is an acceptable frame rate. Our proposed model is called
  xYOLO and exploits the domain specific attributes such as the requirement of
  two classes (ball and goal post), simple shape features of the objects and
  clearly differentiable objects from the background. This allows our model to
  achieve real-time object detection speed with reasonable accuracy.

  In Section~\ref{sec:related-work} we give a brief overview of efforts prior
  to the implementation of the neural networks in this domain, why these
  approaches are not appropriate for the updated environment and the current
  approaches for object detection. In Section~\ref{sec:network} we describe our
  network architecture for xYOLO and how it differs from similar existing work.
  For Section~\ref{sec:experiments} we describe experiments and analyse results
  for comparable networks. Finally, in Section~\ref{sec:discussion} we evaluate
  our work and discuss future work.

\section{Related Work}
\label{sec:related-work}

  Traditionally, in the context of the RoboCup humanoid robotics competition,
  colour segmentation based techniques have been used to detect features of the
  soccer field, such as goals and balls
  \cite{MenasheKelleGenterHannaLiebmanNarvekarZhangStone2017}
  \cite{PolceanuHarrouetBuche2018}. These techniques are fast and can achieve
  good accuracy in simplistic environments, for example the use of an orange
  ball, controlled indoor lighting and yellow coloured goals. However, in light
  of RoboCup's 2050 goal, teams have seen the introduction
  of natural lighting conditions (exposure to sunlight), white goal posts with
  white backgrounds and FIFA balls with a variety of colours. Colour
  segmentation based techniques fail to perform in these challenging scenarios
  and has mostly pushed the competition towards implementing a variety of
  neural network approaches \cite{LeivaCruzBuguenoRuizDelSolar2018}
  \cite{DijkScheunemann2018}.

  CNN based models have shown great progress in terms of object detection
  accuracy in complex scenarios \cite{RedmondFarhadi2018} \cite{Liu_2016},
  \cite{NIPS2012_4824} \cite{googleNet43022} \cite{Redmon2015YouOL}
  \cite{AlbaniYoussefSurianiNardiBloisi2016}. However, these high performing
  computer vision systems based on CNNs, although much leaner than fully
  connected networks, are still both considerably memory and computationally
  exhausting, and achieve real-time performance only on high-end GPU devices.
  For this reason, most of these models are not suitable for low-end devices
  such as smart phones or mobile robots. This limits their use in real-time
  applications such as autonomous humanoid robots playing soccer, as there are
  power and weight considerations. Thus, the development of lightweight,
  computationally efficient models that allow CNNs to work using less memory
  and on minimal computational resources is an active research area
  \cite{CruzLobosTsunekawaRuizDelSolar2017}
  \cite{AlbaniYoussefSurianiNardiBloisi2016}
  \cite{GabelHeuerSchieringGerndt2018} \cite{SpeckBestmannBarros2018}.

  Recently, a large number of research papers have been published on the topic of
  lightweight deep learning models for object detection that are suitable for
  low-end hardware devices \cite{CruzLobosTsunekawaRuizDelSolar2017}
  \cite{SpeckBarrosWeberWermter2016} \cite{CruzTsunekawaSolar2017}
  \cite{li2018tiny} \cite{rastegari2016xnor} \cite{RedmondFarhadi2018}
  \cite{GabelHeuerSchieringGerndt2018} \cite{SpeckBestmannBarros2018}. Most of
  these models are based on SSD \cite{Liu_2016}, SqueezeNet
  \cite{i2016squeezenet}, AlexNet \cite{NIPS2012_4824}, and GoogLeNet
  \cite{googleNet43022}. Generally, in these models the object detection pipeline
  contains several components such as pre-processing, large numbers of
  convolution layers, and post-processing. Classifiers are evaluated at various
  locations in images and at multiple scales using a sliding window approach or
  region proposal methods. These complex object detection pipelines are
  computationally intensive and consequently slow. XNOR-Networks
  \cite{rastegari2016xnor} approximate convolutions using binary operation,
  which is computationally efficient compared to the floats used in traditional
  convolutions. An obvious downside of XNOR networks is the reduction in
  accuracy for similarly sized networks.

  On the other-hand, in \emph{you only look once} (YOLO), object detection is
  framed as a single regression problem. YOLO works at the bounding box level
  rather than pixel level, i.e. YOLO simultaneously predicts bounding boxes and
  associated class probabilities from the entire image in one ``look". One of
  the key advantages of YOLO is its ability to encode contextual information, and as
  a result it makes less mistakes in confusing background patches in an image
  for objects \cite{Redmon2015YouOL} \cite{RedmondFarhadi2018}.

  The ``lighter" version of YOLO v3 \cite{RedmondFarhadi2018}, called
  Tiny-YOLO, was designed with speed in mind and is generally reported as one
  of the better performing models in-terms of speed and accuracy trade-off
  \cite{RedmondFarhadi2018}. Tiny-YOLO has nine convolutional layers and two
  fully connected layers. Our experiments suggest that Tiny-YOLO is able to
  achieve 0.14 FPS on Raspberry Pi 3, which is far from real-time object
  detection.

  From the results reported in \cite{RedmondFarhadi2018}, it can be concluded
  that these object detectors are not able to give real-time performance on
  low-end hardware with minimal computing resources (e.g. humanoid robots with
  a Raspberry Pi as the computing resource). In our robots, we are using one
  compute resource for several different processes, such as the walk engine,
  self-localization, etc. The vision system is left with approximately a single
  core to perform all object detection.

\section{Network}
\label{sec:network}

  \begin{figure*}[]
    \centering
    \noindent\resizebox{\textwidth}{!}{
  \begin{tikzpicture}
    \draw[use as bounding box, transparent] (-1.8,-1.8) rectangle (17.6, 3.2);
    \newcommand{\networkLayer}[6]{
      \def\a{.01 * #1} 
      \def\b{0.02}
      \def\c{0.025 * #2} 
      \def\t{#3} 
      \def\d{#4} 
      \draw[line width=0.3mm](\c+\t,0,\d) -- (\c+\t,\a,\d) -- (\t,\a,\d);                                                      
      \draw[line width=0.3mm](\t,0,\a+\d) -- (\c+\t,0,\a+\d) node[midway,below] {#6} -- (\c+\t,\a,\a+\d) -- (\t,\a,\a+\d) -- (\t,0,\a+\d); 
      \draw[line width=0.3mm](\c+\t,0,\d) -- (\c+\t,0,\a+\d);
      \draw[line width=0.3mm](\c+\t,\a,\d) -- (\c+\t,\a,\a+\d);
      \draw[line width=0.3mm](\t,\a,\d) -- (\t,\a,\a+\d);
      \filldraw[#5] (\t+\b,\b,\a+\d) -- (\c+\t-\b,\b,\a+\d) -- (\c+\t-\b,\a-\b,\a+\d) -- (\t+\b,\a-\b,\a+\d) -- (\t+\b,\b,\a+\d); 
      \filldraw[#5] (\t+\b,\a,\a-\b+\d) -- (\c+\t-\b,\a,\a-\b+\d) -- (\c+\t-\b,\a,\b+\d) -- (\t+\b,\a,\b+\d);
      \ifthenelse {\equal{#5} {}}
      {} 
      {\filldraw[#5] (\c+\t,\b,\a-\b+\d) -- (\c+\t,\b,\b+\d) -- (\c+\t,\a-\b,\b+\d) -- (\c+\t,\a-\b,\a-\b+\d);} 
    }
    \networkLayer{256 * 1.0}{  3}{ 0.2 - 1}{0}{color=blue!40  }{a} 
    \networkLayer{256 * 1.0}{  2}{ 0.4 - 1}{0}{color=orange!40}{b} 
    \networkLayer{128 * 1.5}{  2}{ 0.6 - 1}{0}{color=blue!40  }{c} 
    \networkLayer{128 * 1.5}{  4}{ 0.8 - 1}{0}{color=orange!40}{d} 
    \networkLayer{ 64 * 2.0}{  4}{ 1.0 - 1}{0}{color=blue!40  }{e} 
    \networkLayer{ 64 * 2.0}{  8}{ 1.2 - 1}{0}{color=orange!40}{f} 
    \networkLayer{ 32 * 2.5}{  8}{ 1.4 - 1}{0}{color=blue!40  }{g} 
    \networkLayer{ 32 * 2.5}{ 16}{ 1.7 - 1}{0}{color=orange!40}{h} 
    \networkLayer{ 16 * 3.5}{ 16}{ 2.1 - 1}{0}{color=blue!40  }{i} 
    \networkLayer{ 16 * 3.5}{ 32}{ 2.6 - 1}{0}{color=orange!40}{j} 
    \networkLayer{  8 * 4.5}{ 32}{ 3.5 - 1}{0}{color=blue!40  }{k} 
    \networkLayer{  8 * 4.5}{ 64}{ 4.4 - 1}{0}{color=orange!40}{l} 
    \networkLayer{  8 * 4.5}{ 64}{ 6.1 - 1}{0}{color=cyan!40  }{m} 
    \networkLayer{  8 * 4.5}{128}{ 7.8 - 1}{0}{color=cyan!40  }{n} 
    \networkLayer{  8 * 4.5}{ 32}{11.1 - 1}{0}{color=cyan!40  }{o} 
    \networkLayer{  8 * 4.5}{ 64}{12.0 - 1}{0}{color=cyan!40  }{p} 
    \networkLayer{  4 * 7.5}{  1}{13.8 - 1}{0}{color=pink!40  }{q} 
    \networkLayer{  4 * 7.5}{  1}{14.1 - 1}{0}{color=white    }{r} 
    \networkLayer{  8 * 4.5}{ 32}{14.4 - 1}{0}{color=blue!40  }{s} 
    \networkLayer{  8 * 4.5}{ 16}{15.3 - 1}{0}{color=yellow!40}{t} 
    \networkLayer{  4 * 7.5}{  1}{15.9 - 1}{0}{color=white    }{u} 
    \networkLayer{ 16 * 3.5}{ 48}{16.3 - 1}{0}{color=blue!40  }{v} 
    \networkLayer{ 16 * 3.5}{ 32}{17.6 - 1}{0}{color=blue!40  }{w} 
    \networkLayer{  4 * 7.5}{  1}{18.5 - 1}{0}{color=pink!40  }{x} 
  \end{tikzpicture}
}
    \caption{
      The network structure is as follows:
      \colorbox{blue!40}{convolutional:} a, c, e, g, i, k, s, v, w,
      \colorbox{orange!40}{max pool:} b, d, f, h, j, l,
      \colorbox{cyan!40}{convolutional XNOR:} m, n, o, p,
      \colorbox{pink!40}{yolo:} q, x,
      \colorbox{yellow!40}{upsample:} t,
      \colorbox{white}{route:} r, u.
    }
    \label{fig:network}
  \end{figure*}
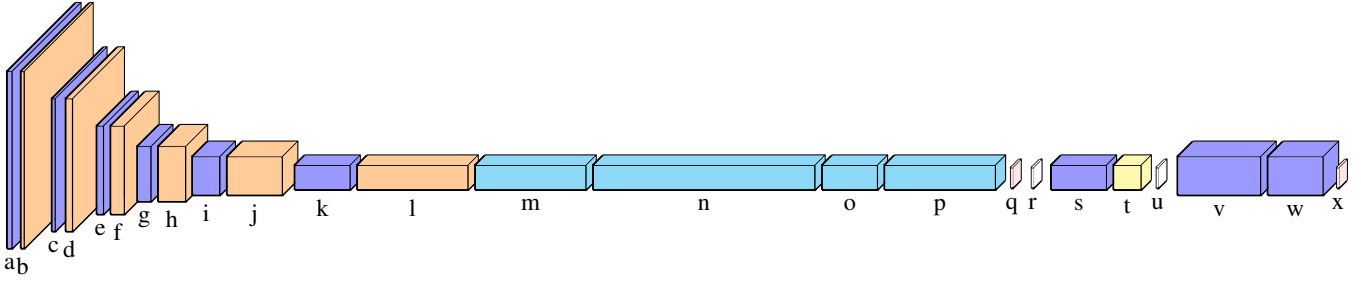

\begin{figure}[tb]
    \centering
    \includegraphics[width=.8\linewidth]{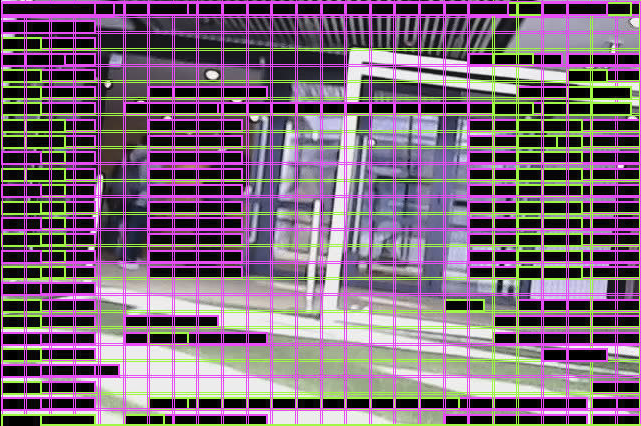}
    \caption{Example of a network that is too small to detect objects.}
    \label{fig:small-net}
  \end{figure}

  Our proposed network, xYOLO, is derived from \emph{YOLO v3 tiny}
  \cite{RedmondFarhadi2018}, specifically we use AlexeyAB's Darknet fork that
  allows for XNOR layers and building on the Raspberry Pi \footnote{Darknet
  fork: \url{https://github.com/AlexeyAB/darknet}}. As seen in
  Figure~\ref{fig:network}, xYOLO utilizes both normal convolutional and XNOR
  layers in both training and recall. The network has several key changes:

  \begin{itemize}
    \item \emph{Reduction in input layer size:} Scaling the input image to 256
      x 256 pixels was the smallest input we could create without sacrificing
      the network's ability to see details at far-distance in the 640 x 480
      original image. Due to limitations of the framework implementation,
      preserving aspect ratio was not easily possible. Switching from RGB to
      grey scale input (three channels down to one) had very little impact on
      speed, but largely affected detection quality, hence we use full colour
      information. Through experimentation we realized that ball detection
      relies on its context, in this case the green field background.
    \item \emph{Heavily reduced number of filters:} Generally, the objects we
      are attempting to detect are quite simple in shape and features, meaning
      this domain specific reduction can be made. We were able to heavily
      reduce the size of the network with this change and increase detection
      speed dramatically.
    \item \emph{Layers $m$, $n$, $o$ and $p$ use XNOR:} Through experimentation
      we found that this part of the network was able to switch to XNOR
      without affecting training or prediction. Whilst the network size remains
      the same, not using floating point arithmetic gave a marginal speed
      increase during detection. When using XNOR throughout the network
      (specifically the convolutional layers between a and k) we found the
      network was unable to detect objects to any accuracy (see
      Figure~\ref{fig:small-net}). We believe the early feature formation in
      the network to be highly important in training and object detection for
      small networks.
  \end{itemize}

  Each year of the RoboCup competition introduces new challenges, where models
  have to be retrained using images collected and
  labelled during the setup time of the competition. Consequently, our approach
  towards designing this network was to reduce the training time to below 45
  minutes, allowing for relatively rapid testing of different network
  configurations and new soccer field conditions. Figure~\ref{fig:small-net} is
  an example of a network where the parameters are reduced too far, such that
  it is incapable of detecting objects. In Figure~\ref{fig:loss} this would
  manifest itself as the loss mean square error not reducing below 6 before
  1,000 iterations or models that are not able to reduce their loss to an
  acceptable value, i.e. below 1.5. Generally we are able to conclude whether a
  network has a reasonable chance of success within the first 15 minutes of
  training.

  \begin{table}[tb]
    \centering
    \begin{tabular}{|c|c|c|c|c|}
      \hline
      \textbf{ID} & \textbf{Filters} & \textbf{Size} & \textbf{Input}  & \textbf{Output} \\
      \hline
      a           &       2          & 3 x 3 / 1     & 256 x 256 x   3 & 256 x 256 x   2 \\
      b           &                  & 2 x 2 / 2     & 256 x 256 x   2 & 128 x 128 x   2 \\
      c           &       4          & 3 x 3 / 1     & 128 x 128 x   2 & 128 x 128 x   4 \\
      d           &                  & 2 x 2 / 2     & 128 x 128 x   4 &  64 x  64 x   4 \\
      e           &       8          & 3 x 3 / 1     &  64 x  64 x   4 &  64 x  64 x   8 \\
      f           &                  & 2 x 2 / 2     &  64 x  64 x   8 &  32 x  32 x   8 \\
      g           &      16          & 3 x 3 / 1     &  32 x  32 x   8 &  32 x  32 x  16 \\
      h           &                  & 2 x 2 / 2     &  32 x  32 x  16 &  16 x  16 x  16 \\
      i           &      32          & 3 x 3 / 1     &  16 x  16 x  16 &  16 x  16 x  32 \\
      j           &                  & 2 x 2 / 2     &  16 x  16 x  32 &   8 x   8 x  32 \\
      k           &      64          & 3 x 3 / 1     &   8 x   8 x  32 &   8 x   8 x  64 \\
      l           &                  & 2 x 2 / 1     &   8 x   8 x  64 &   8 x   8 x  64 \\
      m           &     128          & 3 x 3 / 1     &   8 x   8 x  64 &   8 x   8 x 128 \\
      n           &      32          & 1 x 1 / 1     &   8 x   8 x 128 &   8 x   8 x  32 \\
      o           &      64          & 3 x 3 / 1     &   8 x   8 x  32 &   8 x   8 x  64 \\
      p           &      21          & 1 x 1 / 1     &   8 x   8 x  64 &   8 x   8 x  21 \\
      s           &      16          & 1 x 1 / 1     &   8 x   8 x  32 &   8 x   8 x  16 \\
      t           &                  &        2x     &   8 x   8 x  16 &  16 x  16 x  16 \\
      v           &      32          & 3 x 3 / 1     &  16 x  16 x  48 &  16 x  16 x  32 \\
      w           &      21          & 1 x 1 / 1     &  16 x  16 x  32 &  16 x  16 x  21 \\
      \hline
    \end{tabular}
    \caption{The following is the network structure of xYOLO.}
    \label{tab:network-structure}
  \end{table}

\section{Experiments and Results}
\label{sec:experiments}

  Experiments are conducted using Darknet \cite{darknet13}, an open source
  neural network framework. Darknet is fast (written in C language with many
  optimizations), easy to compile on the Raspberry Pi and supports both CPU and
  GPU training and detection. Our proposed model (xYOLO) is compared against
  Tiny-YOLO (v3) \cite{RedmondFarhadi2018} and Tiny-YOLO-XNOR (v3)
  \cite{rastegari2016xnor}. Tiny-YOLO is reported as a CNN model with good
  trade-off between computational efficiency and object detection accuracy
  \cite{li2018tiny} \cite{JavadiAzarAzamiGhidarySadeghnejadBaltes2017}.
  Tiny-YOLO-XNOR is a lightweight implementation of Tiny-YOLO with XNOR
  \cite{rastegari2016xnor} in the Darknet framework. Each of the models is
  adapted for the use of two classes by adjusting the number of filters before
  the YOLO layers to the following (we have two classes, \emph{ball} and
  \emph{goal}):

  \begin{equation}
    filters = (classes + 5) \times 3
    \label{equ:filter-size}
  \end{equation}

  The models are evaluated on our dataset. The details of the dataset are
  described in Section~\ref{subsec:dataset}. All models were trained using 90\%
  of images in the dataset and 10\% of these images were used during testing.
  Object detection accuracy is measured by mean Average Precision (mAP) and
  F-Score \cite{Lin2014MicrosoftCC}. All of these models are evaluated using
  the default parameters settings. Computational efficiency of the models is
  measured by inference time and train time (minutes). Models are evaluated on
  the Raspberry Pi 3 B, a standard desktop CPU (Intel i7-6700HQ) and a standard
  GPU (Nvidea GTX 960M) environment to measure inference time. Since memory is
  also an issue on low-end hardware, we also compared models using network size
  MB (Mega Bytes) and computational size BFLOPs (Billion FLOPs) performance
  metrics \cite{RedmondFarhadi2018}.

  \subsection{Dataset}
  \label{subsec:dataset}

    One of the contributions of this study is our annotated dataset from the
    RoboCup 2019 competition using cameras mounted on the robots in both the
    controlled and natural lighting scenarios. We also used some images from
    previous competitions via the Image Tagger community-driven project
    \cite{imagetagger2018}. Each of these raw images are manually annotated.
    There are two classes in the dataset: ball and goal post. Traditionally
    people used complete goals as a single object. The inside of the goal is
    hollow and usually only part of the goal on the field is in the camera
    frame, making the detection of a full goal often difficult. In RoboCup,
    generally the ball stays on the ground, thus the robot rarely needs to look
    upwards and detecting only the bottom of the goal posts is ideal. In
    consideration of this, we used bottom of the goal posts to detect goal. In
    this dataset both left and right goal posts are considered as two instances
    of the same class (goal post).

    This dataset contains range of challenging scenarios, such as natural
    lighting (sun light spots on the field), shadows, and blurred images since
    robots are moving. Some of the glimpses of the complexities of the dataset
    can be seen in Figure~\ref{fig:qualitative}. We have open sourced this
    dataset and is available for public use \footnote{Dataset released under a
    Creative Commons license (free login required for access):
    \url{https://imagetagger.bit-bots.de/images/imageset/689/}}.

  \subsection{Comparative Computational Speed}
  \label{subsec:results:comparitive-computational-speed}

    The key focus of this paper is to achieve real-time object detection and
    localization performance on low-end computing hardware such as Raspberry Pi
    3 B. To time models training duration, we used a cloud instance with an
    Nvidea K80 GPU and 55GB RAM. All models were trained for 6000 iterations
    and tested for inference speed in both a standard desktop environment and a
    Raspberry Pi. As shown in Table~\ref{tab:speed}, train times are reported
    on the cloud instance GPU and inference speed is reported on multiple
    hardware targets.

    As shown in the Table~\ref{tab:speed}, xYOLO achieved superior performance
    in terms of computational efficiency compared to the other tested models.
    For train time xYOLO is $\approx$5 times faster than the other two models.
    For inference speed, xYOLO achieved 706.36 FPS on the GPU, which is
    $\approx$7 times faster than Tiny-YOLO-XNOR and $\approx$9 times faster
    than Tiny-YOLO. On desktop CPU, xYOLO performed 87 and 35 times better than
    Tiny-YOLO and Tiny-YOLO-XNOR respectively. On the Raspberry Pi, xYOLO
    performed 69 times faster than Tiny-YOLO and 25 times faster than
    Tiny-YOLO-XNOR. The improved speed gain is due to small input and filters
    size. Tiny-YOLO-XNOR averaged 2.52 FPS on the Raspberry Pi and Tiny-YOLO at
    1 FPS. Both of these models are too slow to be used effectively in the
    RoboCup competition, as games develop quickly. On the other hand, xYOLO is
    capable of $\approx$10 fps on the Raspberry Pi, which is reasonable object
    detection speed, especially for the purpose of humanoid league soccer
    matches.

    In terms of network size, xYOLO is 45 times smaller than Tiny-YOLO and
    $\approx$15 times smaller than Tiny-YOLO-XNOR in network size. Similarly,
    xYOLO requires  0.039 BFLOPs, which is significantly lower than other two
    models. In short, xYOLO outperformed other models on all computational
    efficiency metrics.

    \begin{table}[tb]
      \begin{center}
        \begin{tabular}{|l|c|c|c|c|}
          \hline
          \textbf{Network}              & Tiny-YOLO & Tiny-YOLO-XNOR & xYOLO          \\
          \hline
          \textbf{Train Time (minutes)} & 183       & 174            & \textbf{39}    \\
          \hline
          \textbf{Inference (FPS)}      & 0.89      & 2.52           & 70.69          \\
          \textbf{i7-6700HQ} stdev      & 0.059     & 0.026          & 0.0018         \\
          \hline
          \textbf{Inference (FPS)}      & 79.97     & 97.28          & 706.36         \\
          \textbf{GTX 960M} stdev       & 0.00064   & 0.00055        & 0.00023        \\
          \hline
          \textbf{Inference (FPS)}      & 0.14      & 0.39           & \textbf{9.66}  \\
          \textbf{rPi 3 B} stdev        & 0.064     & 0.012          & 0.0012         \\
          \hline
          \textbf{Size (MB)}            & 34.7      & 12.46          & \textbf{0.82}  \\
          \hline
          \textbf{BFLOPs}               & 5.449     & 5.449          & \textbf{0.039} \\
          \hline
        \end{tabular}

      \end{center}
      \caption{
        Computational efficiency and network size results. Note that the
        reported BFLOPs for Tiny-YOLO-XNOR and xYOLO incorrectly calculate the
        XNOR layers as FLOPs.
      }
      \label{tab:speed}
    \end{table}

  \subsection{Comparative Accuracy}
  \label{subsec:results:comparitive-accuracy}

  \begin{figure}[tb]
    \centering
    \includegraphics[width=.99\linewidth]{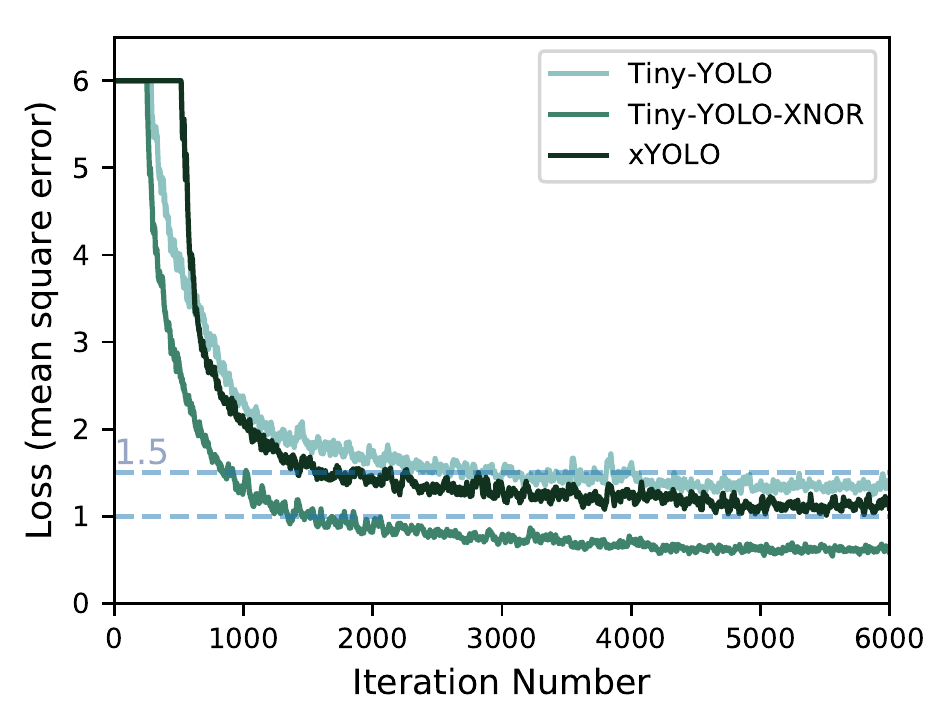}
    \caption{
      Train loss results for the models. It is observed that the models took
      $\approx$4000 iterations to reach their lowest loss.
    }
    \label{fig:loss}
  \end{figure}

  \begin{figure}[tb]
    \centering
    \includegraphics[width=.99\linewidth]{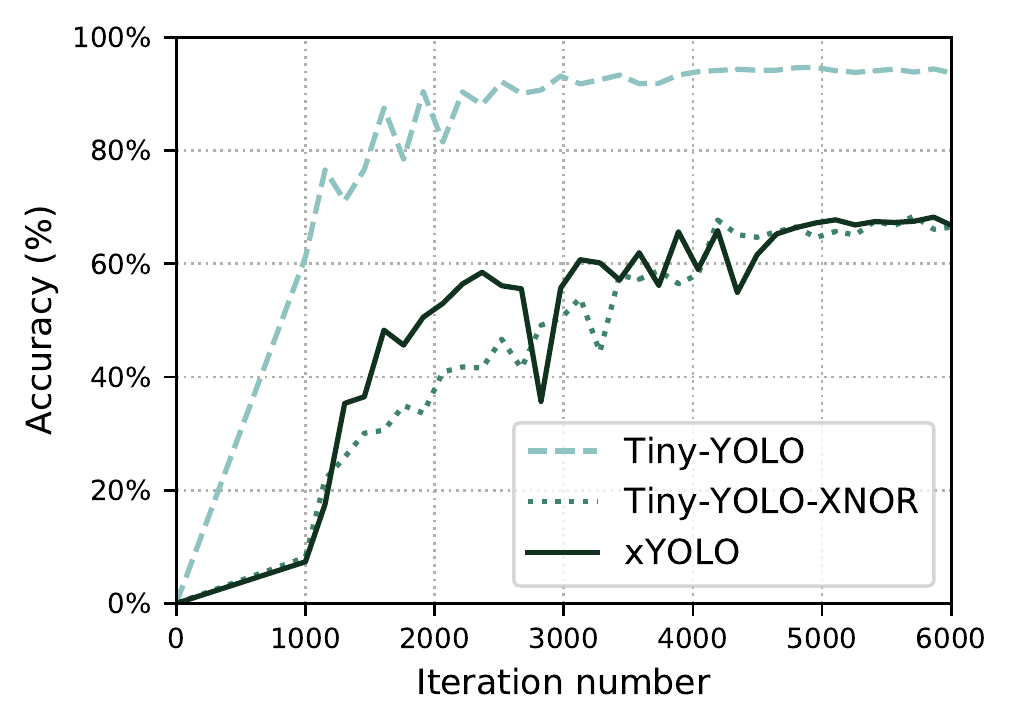}
    \caption{
      Accuracy (mAP) results for the models during training on the validation
      set. It is observed that Tiny-YOLO performed significantly better than
      the other two models.
    }
    \label{fig:map}
  \end{figure}

  \begin{table}[tb]
    \begin{center}
      \begin{tabular}{|l|c|c|c|c|}
        \hline
        \textbf{Algorithm} & \textbf{mAP (Train)} & \textbf{mAP (Test)} & \textbf{F-score} \\
        \hline
        Tiny-YOLO          & 94.65                & 93.69               & 92               \\
        \hline
        Tiny-YOLO-XNOR     & 68.42                & 64.75               & 62               \\
        \hline
        xYOLO              & 68.22                & 66.75               & 68               \\
        \hline
      \end{tabular}
    \end{center}
    \caption{
      Object detection accuracy results on our humanoid soccer dataset (see
      Section~\ref{subsec:dataset}).
    }
    \label{tab:accuracy}
  \end{table}

  The object detection and localization accuracy of the models is measured by
  train loss (mean square error), validation mAP (mean Average Precision on
  validation dataset), inference mAP on test dataset and F-score. We used
  Darknet transfer learning where pre-trained weights
  (\texttt{darknet53.conv.74}) for the Imagenet dataset are used as initial
  weights for training. This transfer learning helps models take less than 1000
  iterations to reduce their loss to less than 6. Figure~\ref{fig:loss} shows
  train loss for the models. Results from the Figure~\ref{fig:loss} suggests
  that the models took $\approx$4000 iterations to stabilized and after that
  loss was not greatly reduced. Tiny-YOLO was able reduce its loss to
  $\approx$0.5, with xYOLO to $\approx$1 and Tiny-YOLO-XNOR to $\approx$1.5.
  Accuracy results (mAP) on the validation set are presented in
  Figure~\ref{fig:map} and Table~\ref{tab:accuracy}. It is observed that models
  have achieved similar accuracy on both train and unseen test sets. Tiny-YOLO
  achieved significantly better object detection accuracy compared to other
  models. xYOLO was able to achieve $\approx$68\% accuracy on validation dataset,
  and $\approx$67\% on the test set, which is good when the speed and size of
  xYOLO is considered.

  \subsection{Qualitative Evaluation}
  \label{subsec:results:qualitative-evaluation}

    \begin{figure*}[!tb]
      \begin{subfigure}{.99\textwidth}
      \centering
        \includegraphics[width=.32\linewidth]{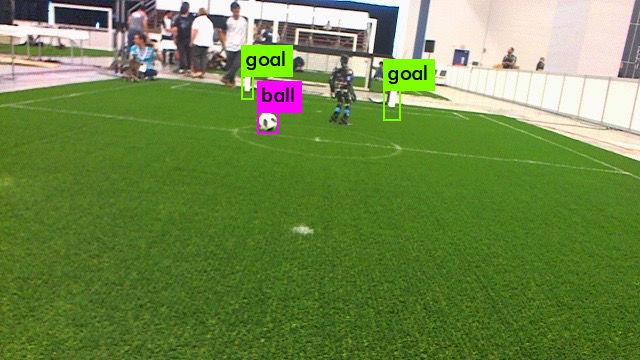}
        \fbox{\includegraphics[width=.32\linewidth]{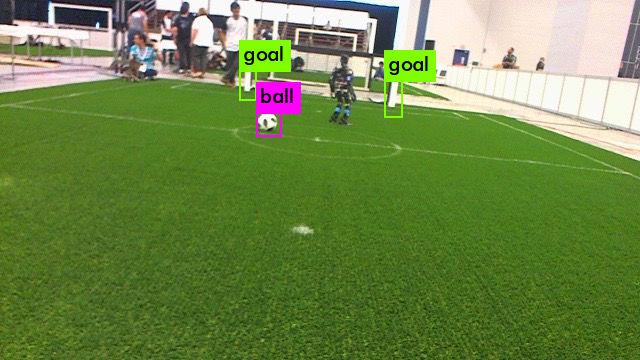}}
        \includegraphics[width=.32\linewidth]{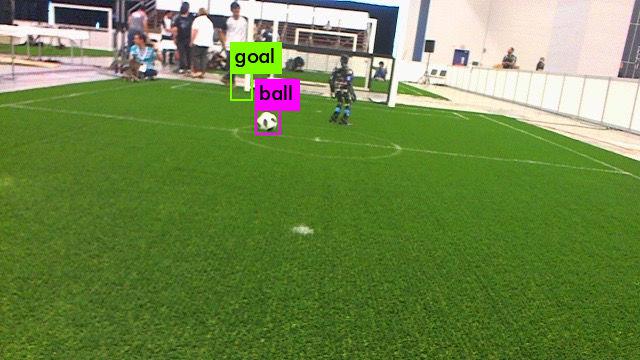}
        \caption{
          Easy detection scenario, ball and goal posts are clear.
        }
        \label{fig:normal_xyolo}
      \end{subfigure}
      \begin{subfigure}{.99\textwidth}
        \centering
        \includegraphics[width=.32\linewidth]{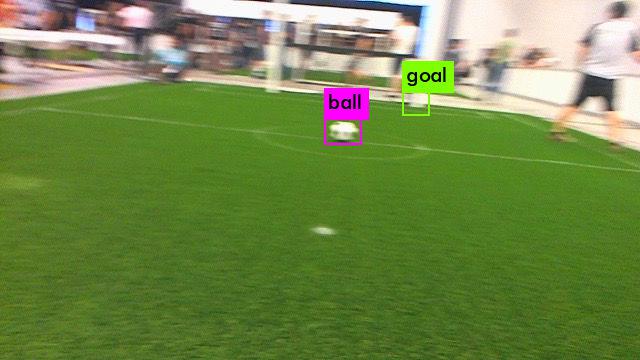}
        \fbox{\includegraphics[width=.32\linewidth]{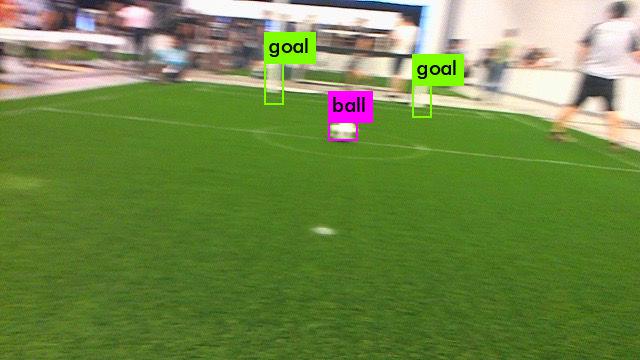}}
        \includegraphics[width=.32\linewidth]{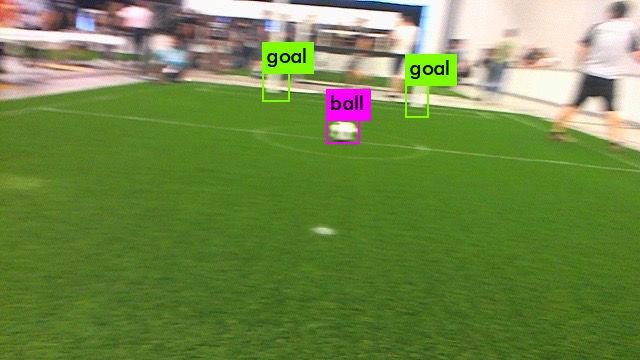}
        \caption{
          As robot's camera is often moving and as a result models have to
          perform in blurred image scenario.
        }
        \label{fig:blur_xyolo}
      \end{subfigure}
      \begin{subfigure}{.99\textwidth}
        \centering
        \includegraphics[width=.32\linewidth]{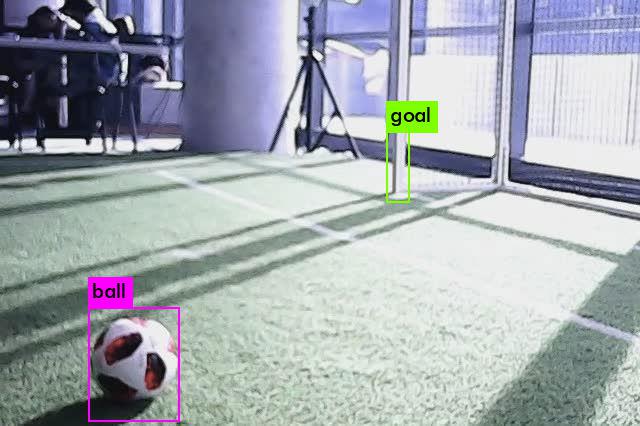}
        \fbox{\includegraphics[width=.32\linewidth]{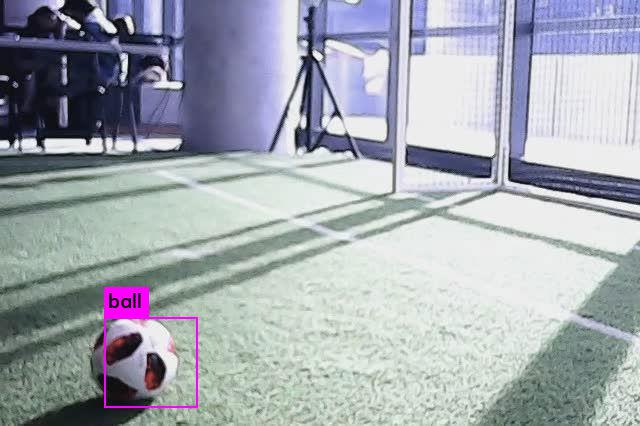}}
        \includegraphics[width=.32\linewidth]{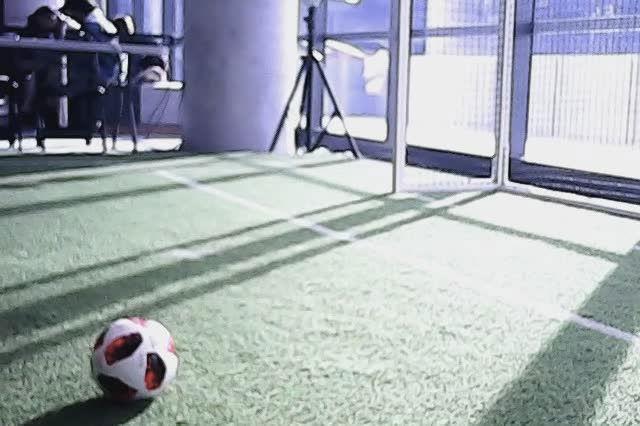}
        \caption{
          Object detection results in natural lighting scenario with shadows on the field.
        }
        \label{fig:shadow_xyolo}
      \end{subfigure}
      \begin{subfigure}{.99\textwidth}
        \centering
        \includegraphics[width=.32\linewidth]{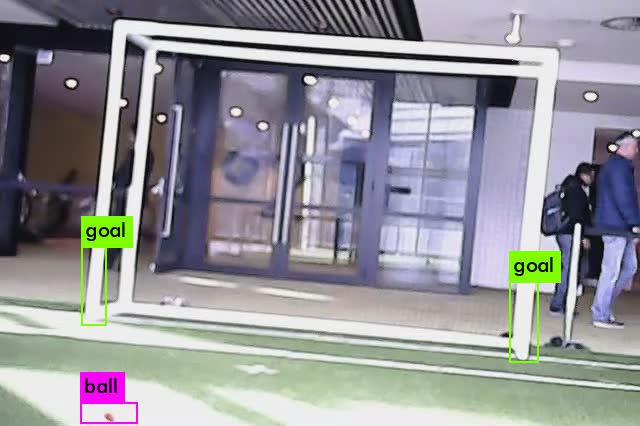}
        \fbox{\includegraphics[width=.32\linewidth]{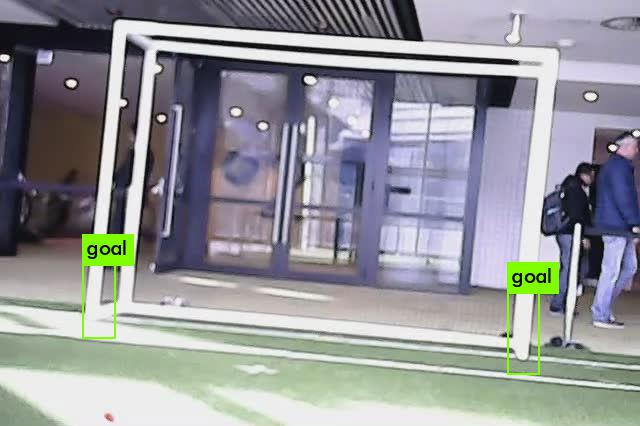}}
        \includegraphics[width=.32\linewidth]{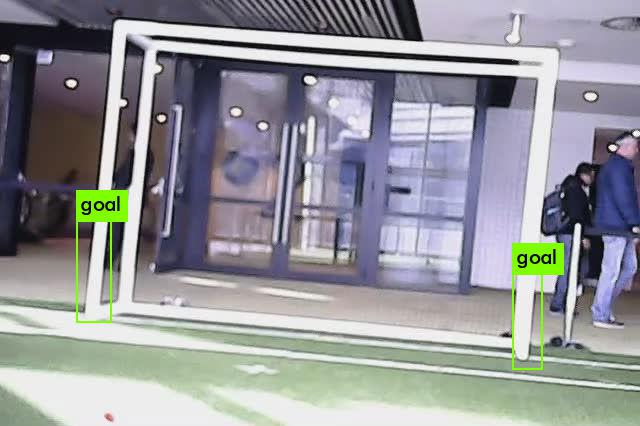}
        \caption{
          Object detection results in natural lighting scenarios with strong
          sunlight spots on the field.
        }
        \label{fig:light_xyoloy}
      \end{subfigure}
      \caption{ Example object detection results produced by the models.
        Left side: Tiny-YOLO \cite{RedmondFarhadi2018}, center: xYOLO,
        right side: Tiny-YOLO-XNOR. \colorbox{magenta!60}{Balls}
        and \colorbox{green!60}{goals} are labelled when each network
        identifies an object that reaches the detection threshold.
        It can be observed that xYOLO has better object detection
        results than Tiny-YOLO-XNOR \cite{rastegari2016xnor} and comparable
        result to Tiny-YOLO.
      }
      \label{fig:qualitative}
    \end{figure*}

    Figure~\ref{fig:qualitative} demonstrates object detection results by each
    of the models on a challenging scenarios. All models were able to
    detect both ball and goal post in easy scenarios, where objects are quite
    clearly seen. It is observed that the Tiny-YOLO struggled to detect one
    goal post in the blurred image scenarios (Figure~\ref{fig:blur_xyolo}). In
    a scenario where there are shadows on the field
    (Figure~\ref{fig:shadow_xyolo}), Tiny-YOLO-XNOR was not able to detect both
    goal post or ball, whereas xYOLO was able to detect the ball but not the goal post
    and Tiny-YOLO was able to detect both objects. In the natural lighting
    scenario with strong sunlight spots on the field, all models performed well
    with Tiny-YOLO able to detect partially observable ball. In
    summary, both Tiny-YOLO and xYOLO have shown advantages in different
    scenarios.

\section{Discussion}
\label{sec:discussion}

  Although xYOLO has less accuracy, it is the only model tested that was able
  to achieve $\approx$10 FPS with an acceptable $\approx$70\% accuracy, making
  it suitable for low-end hardware real-time detection on the Raspberry Pi. For
  humanoid soccer, robots have to make quick decisions (e.g. to detect a
  rolling ball). For this reason, fast models with slightly lower accuracy work
  better than highly accurate but slower models. xYOLO provides a good speed
  and accuracy compromise for humanoid soccer, which was achieved by reducing
  the training time, thereby reducing experiment time and allowed for us to
  fine tune the network to detect objects within the domain.

  In future work we look towards performing pre-processing techniques on the
  input image to further reduce the size of the network. Additionally we want
  to leverage the high correlation of inter-frame data through the use of
  optical flow.

\section{References}
\label{sec:references}

  \begingroup
    \renewcommand{\section}[2]{}
    \IEEEtriggercmd{\enlargethispage{-1cm}}
    \IEEEtriggeratref{23}
    \bibliography{references-db,references-ms}{}
    \bibliographystyle{plain}
  \endgroup

\end{document}